\documentclass[trsc,nonblindrev]{informs3} % current default for manuscript submission

\OneAndAHalfSpacedXI % current default line spacing
\usepackage{natbib}
 \bibpunct[, ]{(}{)}{,}{a}{}{,}%
 \def\bibfont{\small}%

\usepackage[utf8]{inputenc} % allow utf-8 input
\usepackage[T1]{fontenc}    % use 8-bit T1 fonts
\usepackage{hyperref}       % hyperlinks
\usepackage{url}            % simple URL typesetting
\usepackage{booktabs}       % professional-quality tables
\usepackage{amsfonts}       % blackboard math symbols
\usepackage{nicefrac}       % compact symbols for 1/2, etc.
\usepackage{microtype}      % microtypography
\usepackage{xcolor}         % colors
\usepackage{graphicx}
\usepackage{algorithm}
\usepackage[noend]{algpseudocode}
\usepackage{xspace,mfirstuc,tabulary}
\usepackage{listings}

 \algnewcommand{\parState}[1]{\State%
    \parbox[t]{\dimexpr\linewidth-\algmargin}{\strut\hangindent=\algorithmicindent \hangafter=1 #1\strut}}

\TheoremsNumberedThrough     % Preferred (Theorem 1, Lemma 1, Theorem 2)
\EquationsNumberedThrough    % Default: (1), (2), ...
\begin{document}
%%%%%%%%%%%%%%%%

%\setcounter{page}{1}
%\DOI{10.1287/xxxx.0000.0000}%


%\RUNTITLE{}

% \TITLE{Bundling Information Goods of Decreasing Value}
\TITLE{Learning from Drivers to Tackle the Amazon Last Mile Routing Research Challenge}

\ARTICLEAUTHORS{%
\AUTHOR{Chen Wu\footnote{corresponding author:  \EMAIL{wuc@amazon.com}}, Yin Song, Verdi March, Eden Duthie}
\AFF{Amazon Web Services}
%\AUTHOR{Yin Song}
%\AFF{Amazon Web Services, \EMAIL{yinsong@amazon.com}, \URL{}}
%\AUTHOR{Verdi March}
%\AFF{Amazon Web Services, \EMAIL{marcverd@amazon.com}, \URL{}}
%\AUTHOR{Eden Duthie}
%\AFF{Amazon Web Services, \EMAIL{duthiee@amazon.com}, \URL{}}
} % end of the block

\ABSTRACT{%
The goal of the Amazon Last Mile Routing Research Challenge is to integrate the real-life experience of Amazon drivers into the solution of optimal route planning and optimization. This paper presents our method that tackles this challenge by hierarchically combining machine learning and conventional Traveling Salesperson Problem (TSP) solvers. Our method reaps the benefits from both worlds. On the one hand, our method encodes driver know-how by learning a sequential probability model from historical routes at the zone level, where each zone contains a few parcel stops. It then uses a single step policy iteration method, known as the Rollout algorithm, to generate plausible zone sequences sampled from the learned probability model. On the other hand, our method utilizes proven methods developed in the rich TSP literature to sequence stops within each zone efficiently. The outcome of such a combination appeared to be promising. Our method obtained an evaluation score of $0.0374$, which is comparable to what the top three teams have achieved on the official Challenge leaderboard. Moreover, our learning-based method is applicable to driving routes that may exhibit distinct sequential patterns beyond the scope of this Challenge. The source code of our method is publicly available at this GitHub repository\footnote{ \url{https://github.com/aws-samples/amazon-sagemaker-amazon-routing-challenge-sol}}.

%This paper provides technical details of our method and discuss its operational implications based on the evaluation results.  
 % Enter your abstract
}%

%\KEYWORDS{deterministic inventory theory; infinite linear programming duality; 
%  existence of optimal policies; semi-Markov decision process; cyclic schedule}

%\KEYWORDS{}

\maketitle
%%%%%%%%%%%%%%%%%%%%%%%%%%%%%%%%%%%%%%%%%%%%%%%%%%%%%%%%%%%%%%%%%%%%%%

% Samples of sectioning (and labeling) in TRSC
% NOTE: (1) \section and \subsection do NOT end with a period
%       (2) \subsubsection and lower need end punctuation
%       (3) capitalization is as shown (title style).
%
%\section{Introduction.}\label{intro} %%1.
%\subsection{Duality and the Classical EOQ Problem.}\label{class-EOQ} %% 1.1.
%\subsection{Outline.}\label{outline1} %% 1.2.
%\subsubsection{Cyclic Schedules for the General Deterministic SMDP.}
%  \label{cyclic-schedules} %% 1.2.1
%\section{Problem Description.}\label{problemdescription} %% 2.

% Text of your paper here
\section{Introduction}
The last mile of retail delivery and ecommerce fulfillment represents one of the most expensive operations along the entire supply chain. Every added inefficiency during the last mile delivery should thus be investigated for optimization. Route optimization therefore has become a key aspect of the last mile planning. While decades of research and developments in route optimization have made remarkable progress in solving TSP and Vehicle Routing Problems (VRP), it is still challenging to directly apply these route optimization methods for last mile delivery. This is in part because the objectives (such as cost, time, distance, etc.) that these methods strive to optimize are not necessarily in line with preference and behaviour of delivery drivers, who often face unexpected challenges while executing route plans on the ground. As a result, drivers often deviate from planned routes computed by the Amazon Last Mile team's optimization software \cite{Amzscience}.

To address this issue, the organizer of the Amazon Last Mile Routing Research Challenge (\cite{TheChallenge}, \emph{the Challenge} hereafter) provided a set of historical route sequences taken by experienced Amazon drivers during their last mile deliveries. The goal is to encourage alternative methods that can narrow the gap between conventional route optimization methods and the real-life delivery problems. As stated on the Challenge website (\cite{AboutChallenge}), ``\emph{Leveraging learning-based approaches to understand and anticipate these deviations can lead to more efficient and safer routes, more sustainable last-mile delivery operations, more satisfied drivers, and a higher service quality}". As a result, 229 research teams entered the Challenge, and 48 of them were qualified for the final model submission (\cite{TheChallenge}).

The Challenge provides two datasets. The training data set \cite{TrainSet} has 6,112 routes, and the evaluation set \cite{EvalSet} has 3,052 routes. Each route contains a set of 50 to 250 stops. Each stop has metadata including approximate geo-locations, parcel specifications, customer preferred time windows, expected service times, and zone IDs. Each route in the training set is associated with an \emph{actual sequence} taken by the driver. Each actual sequence is assigned a level of quality grade (i.e., high, medium, and low) by the Challenge organizer. Out of the 6,112 actual sequences, 2,718 are of high quality, 3,292 are medium, and 102 are low. All routes in the evaluation set are of high quality as per \cite{ScoreEval}.

For each route in the evaluation set, the Challenge participant is asked to predict its stop sequence such that the dissimilarity between the predicted sequence and its actual sequence is minimized.  Participants use the training set to develop and validate their methods. To do well in the Challenge, their methods need to incorporate driver experience and skills with respect to route sequences. 

To our best knowledge, most modern TSP solvers such as Concorde (\cite{lenstra2009traveling}) and LKH (\cite{helsgaun2000effective}) do not concern driver behaviour. Therefore, the key contributions of our method include: \textbf{(a)} a trained sequential probability model that captures driver experience and behaviour, \textbf{(b)} a Rollout algorithm that generates new zone sequences exhibiting similar sequential patterns captured by the sequential probability model, and \textbf{(c)} a hierarchical scheme that integrates the probability model, the Rollout algorithm, and the conventional TSP solver (LKH) into a complete working solution for last mile route optimization.

%The paper is organised as follow. Section 2 review related work. Section 3 and 4 provide technical details of our top two methods. Section 5 presents the result and discussion.

\section{Related work}
Recently several machine learning-based methods such as \cite{joshi2020learning,bresson2021transformer,kool2018attention} have been developed to address TSP. In these studies, TSP is reformulated as a deep reinforcement learning problem where the actor's policy network is implemented by Graph Neural Network (GNN) or Transformers \cite{vaswani2017attention} trained using the REINFORCE \cite{williams1992simple} method. The actor outputs a probability distribution from which an action is sampled (greedy, beam search, or stochastic) to decide which stop to visit next in an autoregressive fashion. The output attention weights enable the model to stay agnostic to the sequence length, a nice property of pointer networks \cite{vinyals2015pointer}.

These methods have achieved impressive cost (distance) optimality gaps (e.g. 0.39\% in \cite{bresson2021transformer}) compared to state-of-the-art TSP solvers such as Concorde (\cite{lenstra2009traveling}). However, their models do not learn driver behaviour or preferences from past historical routes. As a result, when we directly applied the pre-trained TSP50 model in \cite{bresson2021transformer}) to the Challenge training set, its dissimilarity score of $0.11$ does not rank on the leaderboard. On the other hand, understanding driving behaviour is important because it has been reported in \cite{li2018learning} that significant discrepancies have been observed between drivers' routes and the planned ones across multiple countries. Many reasons could have caused such deviations. For example, drivers prefer repeating their familiar routes, and relying on their past experience to act in response to real-time road conditions.

Therefore, we initially modified the loss function and training procedures in \cite{bresson2021transformer}) to train a sequential decision model using imitation learning (\cite{hussein2017imitation}). However, those initial efforts did not achieve the same level of result reported in this paper. We suspect the reason is that training deep learning-based methods often requires a large number of samples. For example, \cite{bresson2021transformer} used tens of millions of simulated tours in order to reach a very small optimality gap. In contrast, the Challenge training set only has 6,112 examples.

The winning solution submitted by \cite{cook2021constrained} used constrained local search to obtain an evaluation score of 0.0248 (the first place on the leaderboard). \cite{cook2021constrained} derived zone order and precedence constraints by inspecting historical routes, and integrated these constraints as penalty terms when evaluating newly proposed candidates during $k$-opt moves. This is similar to the method used in LKH3 (\cite{helsgaun2017extension}) in order to extend the TSP Solver LKH (\cite{helsgaun2000effective}) to satisfy VRP constraints.  %To our best knowledge, \cite{cook2021constrained} handcrafted all \emph{zone-based} knowledge into constraints to solve the problem. %while we are exploring machine learning driven models to learn these constraints from historical data.
\begin{figure}[h!]
    \centering
      %\fbox{\rule[-.5cm]{0cm}{4cm} \rule[-.5cm]{4cm}{0cm}}
      \includegraphics[width=0.85\textwidth]{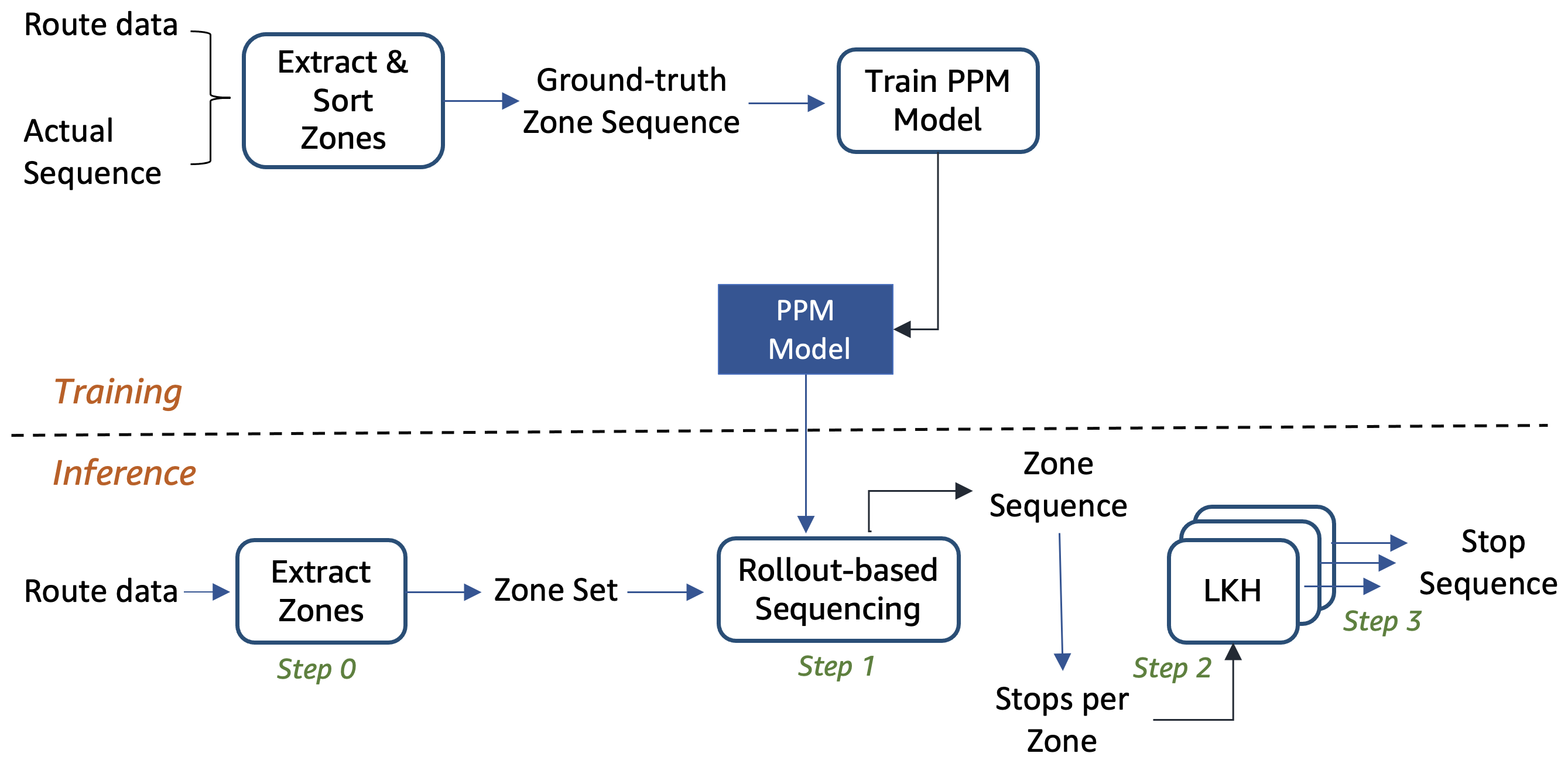}
      \caption{Our method consists of training and inference. Training builds the PPM model to capture zone sequence patterns. Inference uses the PPM model and Rollout-based online search algorithm to produce zone sequences. The LKH TSP solver is then used to sort Stops within each zone.}
      \label{fig:method}
\end{figure}

Our method differs from \cite{cook2021constrained}. First, our method automatically learns a zone-based Markov model from the training set. Instead of observing sequential patterns and encoding them as constraint penalties, we let the high-order Markov model auto-regressively generate new zone sequences guided by the Rollout method for online planning. Since this is completely automated and data-driven, our model is ready to learn continuously emerging patterns (e.g. new drivers joining, new places, new seasons, etc.). Second, our method \emph{generates} synthetic sequence zone in a progressive fashion, where the zone sequence is produced step by step based on sequential decision models developed in the field of reinforcement learning and approximate dynamic programming. This is in contrast to local search methods where the entire solution is iteratively improved via a set of predefined rules or (meta-) heuristics.

\section{Method}
Our method includes the \emph{Training} phase and the \emph{Inference} phase as shown in Figure \ref{fig:method}.
\begin{figure}[h!]
    \centering
      %\fbox{\rule[-.5cm]{0cm}{4cm} \rule[-.5cm]{4cm}{0cm}}
      \includegraphics[width=0.85\textwidth]{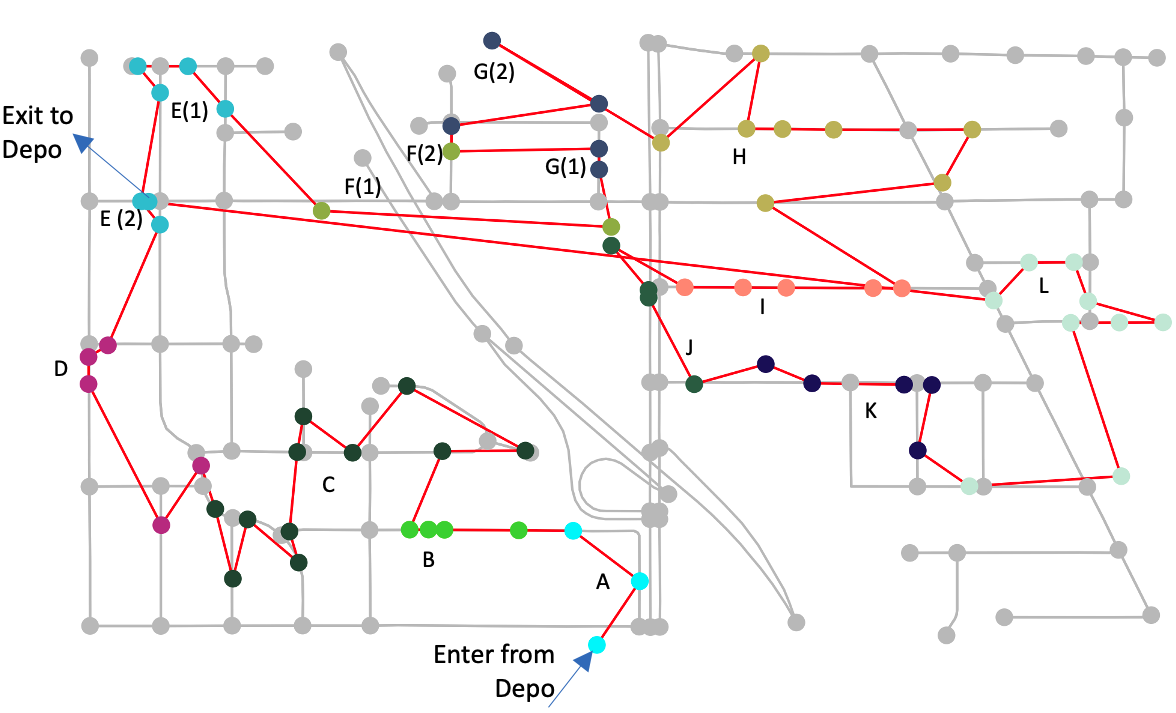}
      \caption{The actual sequence for RouteID\_1a02d09f-2690-4564-87fc-b6d36045e5ba is overlaid on the road network shown as grey edges and nodes. Each delivery stop is colored with its zone ID. The sequence covers the following zones in the order of ``ABCDEFGFGHIJKLE''. Consecutive stops on the actual sequence (connected red edges) tend to share the same zone (color). For zone $E$ and $G$, the number in the bracket indicates the number of times it is being visited. For example, when zone $E$ is visited for the second time, the driver left $E(2)$ for the depo.}
      \label{fig:route_zone}
\end{figure}

\textbf{Training}. We train a Prediction by Partial Matching (PPM) model (\cite{cleary1984data, moffat1990implementing}) using 6,010  \emph{actual sequences} of both high and medium quality. We exclude the 102 poor quality actual sequences because the ground-truth sequences in the evaluation set are all of high quality as per \cite{ScoreEval}. We notice that in \cite{cook2021constrained}, high quality sequences carry more weights than medium ones as constraints. However, we do not differentiate between high and medium quality sequences during training for modelling simplicity and ease of operations. Given an actual sequence of stops, the ground-truth zone sequence is a sequence of zone IDs where each zone ID represents consecutive stops with the same zone ID (e.g. ``C-17.3D'). This is motivated by the fact that stops of the same zone are often visited by drivers in a row as shown in Figure \ref{fig:route_zone}. This is also observed in \cite{cook2021constrained} and others. If a zone ID appears more than once in the ground-truth zone sequence, we keep the one that has the maximum number of stops, which often corresponds to its first appearance, and discard all other appearances. This reduces the ground-truth zone sequence to an approximate ground truth \emph{ZSgt}, which is a lossy representation. For the same example shown in Figure \ref{fig:route_zone}, the ground-truth zone sequence is $ABCDEFGFGHIJKLE$, and the approximate ground-truth \emph{ZSgt} is $ABCDEFGHIJKL$. Figure \ref{fig:route_zone} also suggests that the actual sequence taken by the driver is not necessarily the shortest route with respect to travelling distance and thus deviates from what a conventional TSP solver aims to optimize.

\textbf{Inference}. This phase takes a set of (unordered) stops and generates an ordered sequence of stops using the following steps. \textbf{Step 0} - Derive a set of zones from the set of stops which constitute the “route”. The depo is mapped to a special zone \emph{stz}. \textbf{Step 1} - Sort zones in the most likely order as suggested by the zone sequential model using a Roll-out-based online learning method. \textbf{Step 2} - For each zone, sort its stops using the LKH TSP solver. To incorporate the global perspective, each zone also selectively includes additional stops representing downstream zones. \textbf{Step 3} - Join all zone-sorted stops into the final stop sequence based on the zone sequence produced in Step 1. Details of these steps are discussed in following sub-sections.
% \begin{itemize}
%     \item 
%     \item 
%     \item 
%     \item 
% \end{itemize}
%The following python pseudo-code shows the basic logic of training and inference.
%\begin{minted}{python}
%    # Training
%    ppm_model = train_ppm_model(ground_truth)
%    # Inference
%    for route in all_routes:
%        all_stops = []
%        zone_set = get_zone_set(route) # Step 0
%        sorted_zones = ppm_rollout_sort_zone(zone_set, ppm_model) # Step 1
%        for zone in sorted_zones:
%            sorted_stops = LKH(zone) # Step 2
%            all_stops += sorted_stops # Step 3      
%\end{minted}
\subsection{Why Zone Sorting}
To verify our hypothesis that Step 1 is necessary, we compare the performance of our method with two extreme cases of a sorted zone sequence: (i) the lossy ground truth \emph{ZSgt} vs (ii) re-sort-alphabetically(\emph{ZSgt}). When all the 6,112 routes from the training set are evaluated by the organizer's evaluator, we obtain a score of $0.0116$ for \emph{ZSgt} vs 0.08 for re-sort-alphabetically(\emph{ZSgt}) respectively. This result quantifies experimentally the lower bound (i.e., the best performance) and the upper bound (i.e., the worst possible performance) for our method. Moreover, the gap between the two bounds highlights the importance of Step 1. Note that the close-to-zero score ($0.0116$) produced by \emph{ZSgt} suggests that the lossy ground-truth is practically acceptable to substitute the actual ground-truth.

It is worth noting that the same zone ID maps to different  geo-locations in different routes. For example, zone ``A-1.2D'' include 12 stops that have longitudes around $-118$ in one route. But it corresponds to 11 stops that have longitudes around $-122$ in another route. However, our initial experiments suggest that training location-specific zone models has little effect on training scores. We therefore treat zones with the same name as identical regardless of their corresponding geo-locations.
% \begin{figure}[tb]
%     \centering
%       %\fbox{\rule[-.5cm]{0cm}{4cm} \rule[-.5cm]{4cm}{0cm}}
%       \includegraphics[width=0.85\textwidth]{zone_len_hist}
%       \caption{The distribution of the number of unique zones per route in the training set. Most routes have around 20 or 21 zones. In contrast, the average number of stops per route is around 160.}
%       \label{fig:zone_len_hist}
% \end{figure}

\subsection{Probabilistic Model with PPM}
\begin{figure}[h!]
    \centering
      %\fbox{\rule[-.5cm]{0cm}{4cm} \rule[-.5cm]{4cm}{0cm}}
      \includegraphics[width=0.2\textwidth]{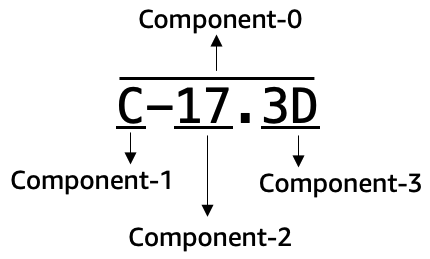}
      \caption{A example of splitting a zone ID into 4 components using a naive string tokenization scheme.}
      \label{fig:zone_component}
\end{figure}

First, we train a model to capture sequential inter-dependency between consecutive zones in \emph{ZSgt}. Specifically, we use the PPM-D method \cite{shkarin2002ppm} to train a high-order Markov model \emph{PM}. During model inference, \emph{PM} estimates the probability of observing a target zone, e.g. 'C-17.2E' after observing a sequential context, such as ['C-17.3D', 'C-17.2D', 'C-17.1D', 'C-17.1E']. \emph{PM} learns this type of conditional probabilities from the training set. One advantage of \emph{PM} is that, by using the ``escape" mechanism, it still works reasonably well even with unknown zone IDs on unseen datasets. We believe this is a necessary ingredient to operationalize our method for production, where misaligned zone IDs are expected to be common. Hence, it is important for \emph{PM} to estimate their probabilities without assuming too much prior knowledge.
%This phenomenon already occurs with the training set and the evaluation set. Both sets share 7,682 zones in common. However, 1,022 zones only appear in the evaluation set but unseen in the training set. 
To support multi-components inside a zone ID, \emph{PM} aggregates conditional probabilities computed separately across four zone components -  the original zone ID, i.e., Component-0 (e.g. C-17.3D), Component-1 (e.g. C), Component-2 (e.g. 17) and Component-3 (e.g. 3D). An example is given in Figure \ref{fig:zone_component}. In comparison to \cite{cook2021constrained}, we do not derive detailed heuristics (e.g. zone hierarchies, re-ordered tokens, etc.) to split the zone. Instead, we tokenize a zone ID with any non-digit or non-letter characters for simplicity. If the zone naming convention changes over time, such a simple strategy that does not overfit current data will be more likely to work. After calculating a sum of conditional probabilities for each zone component, \emph{PM} then computes a weighted average over these four component sums. The weight associated with each component is configured as a hyper-parameter, but has been fixed to 0.25 without any fine-tuning for the Challenge.

\subsection{Sorting Zones with Rollout}
In this section, we describe the Rollout algorithm to generate a zone sequence that maximizes the conditional probabilities summed over the entire sequence. Rollout techniques, first proposed in \cite{tesauro1996line} to beat human players in the Backgammon game, are widely used in many reinforcement learning algorithms to simulate and approximate state value functions.

The optimal cost function $J^*(x)$ is the unique solution of the Bellman equation 
$$J^*(x) = \max_{u\in U}\left[g\big(x,u\big) + J^*\big(F(x, u)\big)\right]\vspace{-0.1in}$$

Here $x$ is the current route \emph{state}, which consists of a set of visited zones thus far. $u$ is the action that determines which next zone the driver needs to visit. $F$ is the route simulator that transitions the route into the next state $x'$ when action $u$ is taken given current state $x$. $g$ is the immediate reward given by the PPM model as the conditional probability. Function $J^*$ is the optimal value function, which returns the expected total reward for a state $x$ if we follow the optimal policy starting from $x$. $J^*$ is expressed by itself given the next state $F(x, u)$, and this recursive structure represents Bellman's principle of optimality \cite{bellman1952theory}. Obtaining the optimal solution of the Bellman equation requires a factorial time complexity, which becomes infeasible in practice even when the number of sequential decision steps $N$ is relatively small, e.g. between $20$ to $50$ as the length of the zone sequence.

We use Rollout (\cite{tesauro1996line, bertsekas2005dynamic}) to obtain a suboptimal solution that provides a good trade-off between solution optimality and execution time complexity. In particular, we replace $J^*$ in the Bellman equation with an approximation $\tilde J$ (aka the baseline policy). The corresponding suboptimal policy $\tilde{\mu}$ is obtained by the one-step lookahead maximization.
\vspace{-5pt}
\begin{equation}
\label{eq:rollout}
\tilde \mu(x)\in\arg\max_{u\in U}\left[ g(x,u)+ \tilde J\big(F(x,u)\big)\right].
\vspace{-0.08in}
\end{equation}
Here $\tilde J$ is a greedy PPM-based baseline policy, which somewhat corresponds to the estimated state value function as the second term of the $Q$-value function. But unlike $Q$-learning (\cite{watkins1992q}), $\tilde J$ is not learned from sampling many trajectories, but obtained by simply following the PPM baseline policy.

Concretely, we define a set of states $(x_1, x_2,...,x_N)$ where $N$ is the number of unique zones in the route. A state $x_k$ represents the first $k$ zones and their order in the zone sequence solution. We consider $x_k$ as a partial solution  $PS_k$ up to step $k$. By symmetry, $x_k$ also corresponds to a set $U$ (Equation \ref{eq:rollout}) of $N - k$ zones that are not yet in $PS_k$ after step $k$. In order to transition from state $k$ to $k+1$, we decide which zone $z_{k+1}$ to visit next, and append $z_{k+1}$ as the last element of $PS_{k+1}$. This is done by applying Equation \ref{eq:rollout} to find the action $\tilde \mu(x_k)$ that leads to zone $z_{k+1}$. While there are up to $N-k$ available zones in $U$ to choose from as $z_{k+1}$, Equation \ref{eq:rollout} only returns $z^*_{k+1}$ that maximizes the sum of the immediate reward $g(x,u_{k})$ and the future reward-to-go $\tilde J$. 

To calculate $g(x,u_{k})$, we query the PPM model to obtain the conditional probability of observing $z_{k+1}$ given the $k$-element zone sequence in the partial solution $PS_k$. For example, if $PS_k$ is a sequence of $CBA$, and the selected $z_{k+1}$ is $D$, the PPM model returns $\text{Probability}(D|CBA)$ as $g(x,u_{k})$. To calculate $\tilde J$, we first use the PPM baseline policy to ``rollout'' a \emph{greedy} zone sequence $ZS_g$ from $U$. $\tilde J$ is then obtained by sliding the PPM model sequentially over $ZS_g$ using a context with a fixed window size (i.e., $5$) and summing up all conditional probabilities. $ZS_g$ is \emph{greedy} because the PPM baseline policy is a greedy way to generate a zone sequence from $U$. At each step $j$, the baseline policy always chooses the next zone $z^*_{j+1}$ such that 
$$z^*_{j+1}\in\arg\max_{z_{j+1}\in U}\left[ \text{Probability}(z_{1}, z_{2},...,z_j|z_{j+1})\right]$$

In summary, at each state $x_k$, we perform one-step lookahead and rollouts for all $N - k$ next zone candidates in $U$, and find the zone that maximizes the sum of the immediate reward and the reward-to-go. We perform such one-step lookahead maximization for each state in $(x_1, x_2,...,x_N)$, starting from $x_1$ until we complete all $N$ states. 

\subsection{Sorting Stops with LKH}
This step leverages an existing TSP solver such as LKH to sort stops within each zone. First, for each zone $zo$ in a sorted zone sequence $ZS$ produced in the last step (Section 3.3), we compute its ``representative” node $rn$, which is essentially the median coordinates over all stops in $zo$. Next, we sequentially iterate through $ZS$. For each zone $zo$ in $ZS$, we extract its stops to form a set $T$. We then add to $T$ representative nodes of all zones that are sorted after $zo$ in $ZS$. This is to inject global perspectives into $T$ so that the TSP solution is not myopic to a local optimum. For the same reason, we also add two special nodes into $T$, the last stop $ls$ in the preceding zone to $zo$ in $ZS$, and the depo. Last, we run the LKH solver on $T$ with the starting node being $ls$. From the sorted stop sequence returned by LKH, we remove all representative nodes, $ls$ and depo, and return the sorted stop sequence $SS$ for $zo$. We repeat this for all zones in $ZS$ and concatenate all $SS$s into a single sorted stop sequence as the final result.

\begin{figure}[h!]
    \centering
      %\fbox{\rule[-.5cm]{0cm}{4cm} \rule[-.5cm]{4cm}{0cm}}
      \includegraphics[width=0.6\textwidth]{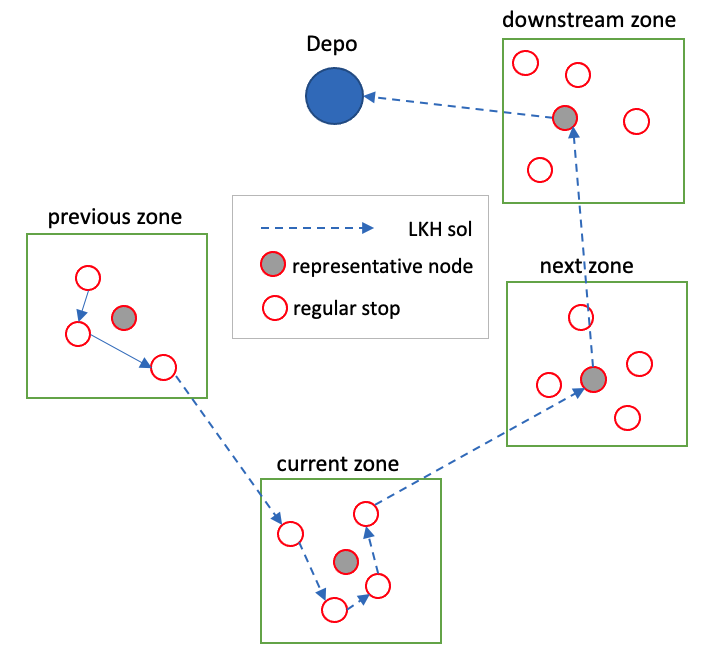}
      \caption{LKH is used to sort stops within each zone. Dotted arrows represent the sequence produced by the LKH TSP solver. The sequence includes 4 stops within the current zone, one stop from the previous zone and two representative nodes for the next zone and the downstream zone respectively.}
      \label{fig:lkh_stop}
\end{figure}

\section{Implementation}
We implement the PPM model and the Rollout algorithm in Python. Sequencing zone-specific stops uses the LKH (version 2.0.9) implementation \cite{helsgaun2000effective} of the Lin-Kernighan TSP heuristic. Since LKH is written in C, we use the Python subprocess module to directly invoke LKH binary executable. We adapt an existing LKH Python wrapper \cite{PyTSP} to auto-generate both the input matrix file and the parameter file. We change the `TYPE' of the distance matrix to ATSP, and implemented a batch version of the wrapper for invoking multiple LKH sub-processes in parallel during Rollout executions. Source code will be made available on the public git repository. %The Git repository will also include a set of Jupyter notebooks as a ``quick-start” tutorial for readers to try our method.

We integrate PPM model inference, Rollout algorithms and LKH (on both training and evaluation set) executions within a single Amazon SageMaker processing job. Since training takes less than 4 seconds, it was done on the SageMaker notebook instance at this stage. Integration of PPM training into a repeatable pipeline is left for our future work.

\section{Results}
Table \ref{table:2} presents the score of our method evaluated against the evaluation set that contains 3,052 routes. The formula used to calculate the score is defined in \cite{ScoreEval} and we used its open source implementation (\cite{ScoreEvalGit}) without any modifications. The evaluation was performed on an AWS ml.m5.4xlarge instance, and it took our method $\approx 1.15$ seconds to sequence a single route using one CPU core on that instance. A quick profiling run suggests that zone sequencing using PPM and Rollout takes around 0.65 seconds, and sequencing stops using LKH takes about 0.5 seconds. The PPM model was trained on the Amazon SageMaker ml.m5.4xlarge notebook instance.

\begin{table}[h!]
    \begin{center}
    \begin{tabular}{@{}lllll@{}}
    \toprule
     & $N$ & \textbf{Score} & \textbf{Training time} & \textbf{Inference time} \\ \midrule
    Training & 6,112 & 0.0313         & \multicolumn{1}{r}{3.7 seconds}         & \multicolumn{1}{r}{1.15 seconds $/$ route}   \\ \midrule
    Evaluation & 3,052 & \textbf{0.0374}         & \multicolumn{1}{r}{}         & \multicolumn{1}{r}{1.15 seconds $/$ route}   \\ \bottomrule
    \end{tabular}
    \end{center}
    \caption{Final scores of our method on the Training and Evaluation Sets.}
    \label{table:2}
    \end{table}
\begin{table}[h!]
    \begin{center}
    \begin{tabular}{@{}lcl@{}}
    \toprule
    \textbf{Team}               & \textbf{Rank} & \textbf{Score}  \\ \midrule
    \emph{Just Passing Through} & 1             & 0.0248                                                     \\
    \emph{Permission Denied}    & 2             & 0.0353                                                       \\
    \emph{Sky is the Limit}     & 3             & 0.0391                                                      \\ 
    \emph{MEGI}     & 4             & 0.0436                                                      \\ 
    \emph{UPB}     & 5             & 0.0484                                                     \\ \bottomrule
    \end{tabular}
    \caption{Scores of the top 5 (out of 24) teams listed on the Challenge leaderboard (\cite{LeaderBoard}). Our evaluation score of \textbf{0.0374} is comparable to what the 3rd-placed team has achieved.}
    \label{table:3}
    \end{center}
    \end{table}

%Our evaluation score of $0.0374$ (Table \ref{table:2}) is comparable to what the 3rd-placed team has achieved in the Challenge (Table \ref{table:3}). 
We commend the outstanding score of $0.0248$ obtained by the top team (\cite{cook2021constrained}). The main difference between \cite{cook2021constrained} and our method boils down to two key questions: (1) who acquires the zone knowledge? and (2) how is it encoded and used? \cite{cook2021constrained} translated their observations of zone order and precedence into 5 to 6 constraints, and encoded these constraints analytically as penalty terms in the objective function of the local search method. We take a different approach where we let the PPM model acquire and encode the zone knowledge by learning higher-order conditional probabilities automatically. We also use the standard string tokenization scheme (Figure \ref{fig:zone_component}) without acquiring any additional knowledge specific to the Challenge. It makes our method adaptive and cost-efficient by shifting the responsibility of making sense of complex patterns to a model that can learn from data automatically. Instead of observing, deriving and fine-tuning heuristics repeatedly, one can apply our method out-of-the-box to unseen routes with driving patterns different from those discovered in this Challenge. Our method is grounded in \emph{maximum likelihood estimation} for optimizing local inter-zone correlations and \emph{principle of optimality} for generating the optimal global sequence.

Table \ref{table:2} shows that it takes $3.7$ seconds to train the PPM model. If we linearly scale the training throughput based on this rate without considering other bottlenecks, it will take one minute to train every 100,000 routes. It is true that training appears to be a one-off task in the Challenge. But for daily operations that continuously ingest datasets from different regions (suburbs, cities, states) at different times (seasons, month of the year, etc.) with constantly changing traffic conditions, it is highly desirable for the last mile operators to re-train the model in order to capture the latest Markov model from the data. In fact, when more new zone patterns accumulate, it is essential to re-build the PPM model frequently. Nevertheless, this could potentially increase the complexity and cost of model deployment for production workloads. For example, we envisage that memory footprint and distributed access and synchronisation of the PPM model could be potential bottlenecks, and thus require further investigations and refactored implementations.

\section{Conclusions}
In this paper, we discussed our method to tackle the Amazon Last Mile Routing Research Challenge. We presented a hierarchical method to combine machine learning (ML) and a conventional TSP solver for effective route optimization. Specifically, we trained an Prediction by Partial Matching (PPM) model that captures higher order sequential patterns at the zone level. We then applied the Rollout algorithm, developed in the field of reinforcement learning, to sample and generate a zone sequence from a given set of zones. The reward (objective) of the generated sequence is to maximize its conditional probabilities as per the learned PPM model. Finally, we used an existing TSP solver, LKH, to sequence stops for each zone at the lower level. The performance of our method is comparable to the top three (out of 24) teams listed on the Challenge leaderboard.

% Acknowledgments here
\ACKNOWLEDGMENT{%
We would like to thank \emph{Josiah Davis} for performing data pre-processing and model training at the initial phase of this work. We would like to thank \emph{Tyler Mullenbach} and \emph{Helmut Katzgraber} for their useful feedback to improve the paper. We would like to thank \emph{Daniel Merchán} from the Amazon Last Mile Science Team for providing us with the dataset and useful information.

% Enter the text of acknowledgments here
}% Leave this (end of acknowledgment)   

% Appendix here
% Options are (1) APPENDIX (with or without general title) or 
%             (2) APPENDICES (if it has more than one unrelated sections)
% Outcomment the appropriate case if necessary
%
% \begin{APPENDIX}{<Title of the Appendix>}
% \end{APPENDIX}
%
%   or 
%
% \begin{APPENDICES}
% \section{<Title of Section A>}
% \section{<Title of Section B>}
% etc
% \end{APPENDICES}

% References here (outcomment the appropriate case) 

% CASE 1: BiBTeX used to constantly update the references 
%   (while the paper is being written).
\bibliographystyle{informs2014trsc} % outcomment this and next line in Case 1
\bibliography{transportation-science-template} % if more than one, comma separated

% CASE 2: BiBTeX used to generate mypaper.bbl (to be further fine tuned)
%\input{mypaper.bbl} % outcomment this line in Case 2

\end{document}